\documentclass[letterpaper]{article} 
\usepackage{aaai25}  
\usepackage{times}  
\usepackage{helvet}  
\usepackage{courier}  
\usepackage[hyphens]{url}  
\usepackage{graphicx} 

\usepackage{nicefrac}       
\usepackage{microtype}      
\usepackage{xcolor,colortbl}         
\usepackage{latexsym}
\usepackage{amssymb}
\usepackage{amsmath}
\usepackage{amsthm}
\usepackage{booktabs}
\usepackage{enumitem}
\usepackage{times}
\usepackage{helvet}
\usepackage{courier}
\usepackage{algorithm2e}
\usepackage{mathtools}
\newtheorem{defi}{Definition}
\newtheorem{theorem}{Theorem}
\newtheorem{prop}{Proposition}
\usepackage{threeparttable}
\usepackage{multirow}
\usepackage{subcaption}
\usepackage{bbding}
\usepackage{array}
\usepackage{pgf}

\DeclareMathOperator*{\argmin}{argmin}

\usepackage{natbib}  
\usepackage{caption} 
\title{FSL-Rectifier: Rectify Outliers in Few-Shot Learning via Test-Time Augmentation}
\author {
    Yunwei Bai\textsuperscript{\rm 1},
    Ying Kiat Tan\textsuperscript{\rm 1},
    Shiming Chen\textsuperscript{\rm 2},
    Yao Shu\textsuperscript{\rm 3},
    Tsuhan Chen\textsuperscript{\rm 1}
}
\affiliations {
    \textsuperscript{\rm 1}National University of Singapore\\
    \textsuperscript{\rm 2}Mohamed bin Zayed University of Artificial Intelligence\\
    \textsuperscript{\rm 3}Guangdong Lab of AI and Digital Economy (SZ)\\
    \{baiyunwei, yingkiat\}@u.nus.edu, shimingchen@gmail.com, shuyao@gml.ac.cn, dprtchen@nus.edu.sg
}

\begin{document}

\maketitle

\begin{abstract}
Few-shot learning (FSL) commonly requires a model to identify images (queries) that belong to classes unseen during training, based on a few labelled samples of the new classes (support set) as reference. So far, plenty of algorithms involve training data augmentation to improve the generalization capability of FSL models, but outlier queries or support images during inference can still pose great generalization challenges.
In this work, to reduce the bias caused by the outlier samples, we generate additional test-class samples by combining original samples with suitable train-class samples via a generative image combiner. Then, we obtain averaged features via an augmentor, which leads to more typical representations through the averaging. We experimentally and theoretically demonstrate the effectiveness of our method, obtaining a test accuracy improvement proportion of around 10\% (e.g., from 46.86\% to 53.28\%) for trained FSL models. Importantly, given a pretrained image combiner, our method is training-free for off-the-shelf FSL models, whose performance can be improved without extra datasets nor further training of the models themselves. Codes are available at \url{https://github.com/WendyBaiYunwei/FSL-Rectifier-Pub}.
\end{abstract}

\begin{figure*}[ht]
\includegraphics[width=\textwidth]{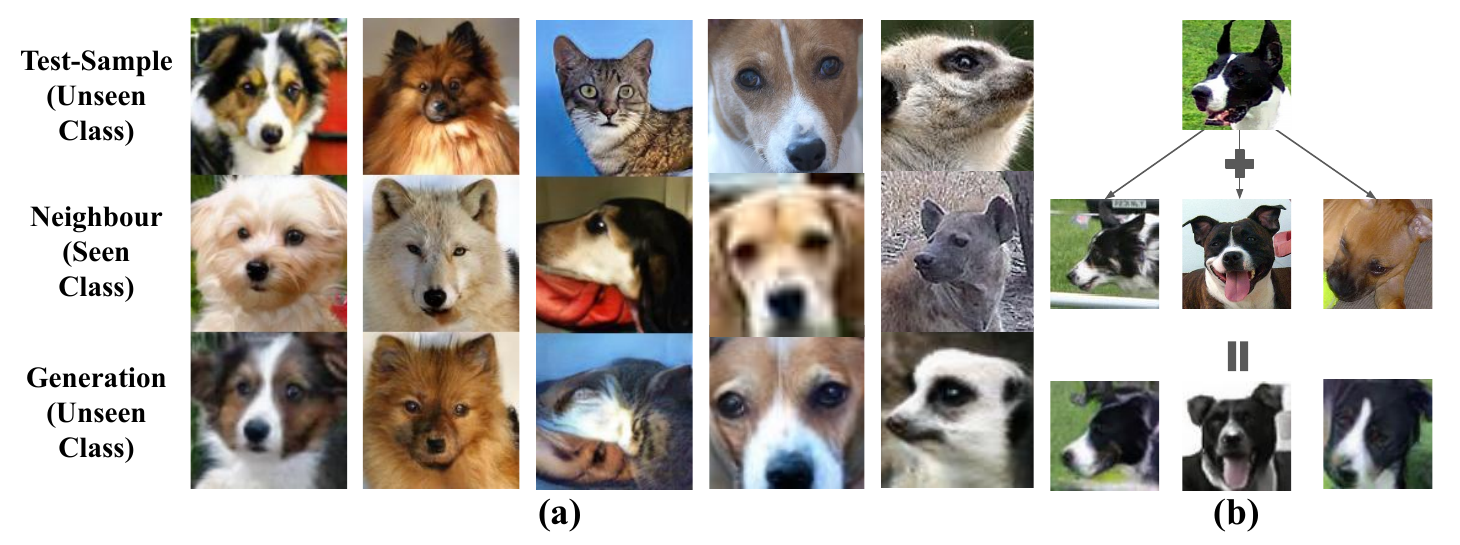}
\caption{Illustration of our key idea. (a): The first image in each column of the animals dataset is the original test sample, the second is \text{neighbour} sample from the test class. The last image is the generation based on style of the original test sample and general shape of the \text{neighbour}. (b): Given different neighbour samples, each test sample can be augmented to become multiple copies.}
\label{translator}
\end{figure*}
\section{Introduction}

Although deep learning has gained much practical success, deep neural networks (DNN) often require a large amount of data to perform well \citep{marcus2018deep}. Usually, there is a high expense for label collection, and some classes of data can be rare and practically difficult to collect \citep{yu2015lsun, marcus2018deep, wang2020generalizing, whang2023data}. For example, in facial expression recognition, dataset class imbalance problems are prevalent, as happy smiley faces are generally easier to collect than facial expressions of disgust \citep{ciubotaru2019revisiting}. Few-Shot Learning (FSL) is a classification problem formulation tackling the issues of limited test-class data. Generally, FSL models can classify unseen test classes when presented with a \textit{support set} comprising only a few labelled test samples as reference~\citep{snell2017prototypical, ye2020fewshot, sung2018learning, wang2020generalizing, dhillon2019baseline}. Nevertheless, while the data constraint, characterized by the absence of test-class data during training, is practically meaningful, it also poses generalization challenges to DNN-based FSL models \citep{wang2020generalizing}. The challenge can be exacerbated by unconventionality of test data samples \citep{wang2020generalizing, kim2019variational, snell2017prototypical}. 

So far, different algorithms are proposed to tackle the issue of high generalization errors in FSL models. Most works involve training augmentation for enhancing the generalization capacity of FSL models \citep{mishra2018generative, verma2018generalized, schwartz2018delta, hariharan2017low, wang2018low, gao2018low}. However, such augmentation involves model training over an increased amount of data, costing extra computation resources over every additional FSL model trained. Furthermore, augmentation over the training dataset can be limited in performance, since the gap between the testing data and the augmented training data may still remain wide.

In this work, we propose to tackle FSL via test-time augmentation instead of training-time augmentation. Our method is named \textit{FSL-Rectifier}, which essentially augments the original test samples before considering the averaged augmentation during classification. For example, given a side-facing wolf test sample, we pick a suitable training sample, say a front-facing dog. Then, we convert the front-facing dog to the wolf class, producing a front-facing wolf image. A trained FSL model then makes classification based on averaged features of both the side-facing wolf and the front-facing wolf, instead of just one single side-facing wolf that may have been an outlier. According to extensive literature, feature averaging can reduce bias, effectively leading to more typical feature representations \citep{zhou2012ensemble, kimura2024understandingtesttimeaugmentation, shanmugam2021better}.

To achieve this, we have three main components: 1) \textit{image combiner}, 2) \textit{neighbour selector} and 3) \textit{augmentor}. For the \textit{image combiner}, we have a generative image translator model pretrained over the training classes. The image combiner considers two images, combining the general shape (e.g., animal pose and position of eyes) of one image and the style (e.g., animal's class-defining fur style) of another via a Generative Adversarial Network (GAN) mechanism \citep{liu2019few}. The combination is illustrated in Figure \ref{translator}(a). During testing of any off-the-shelf FSL models, the \textit{neighbour selector} can select
a better candidate to be combined with a test sample for generation. This test sample can either be a query to be classified, or a support set sample. Finally, through the \textit{augmentor}, we average representations of augmented copies and the original test samples. The augmentation is illustrated in Figure \ref{translator}(b). Therefore, our FSL-Rectifier can make the averaged representations closer to their centroids, correcting certain outlier predictions of an existing FSL model. Besides, our method is test-time-only. Given pretrained image translator, our method can be directly applied to trained FSL models. 

Our contributions include the following:
\begin{itemize}
    \item We propose a novel test-augmentation pipeline for FSL, including an image combiner, a neighbour selector and an augmentor. The image combiner can generate images based on general shape of one image and class-defining features of another. The neighbour selector picks suitable training images whose general shape is to be combined with the class-defining features of original test samples. The augmentor performs feature-averaging across the original test samples and the generated test samples, mitigating the outlier effect in original test samples.
    \item We conduct various experiments and analysis to verify the feasibility of our idea. On a dataset consisting of animal faces, we can achieve around 4\% improvement for a trained FSL model, over the original baseline without any augmentation. The improvement is achieved despite limitation in our image generation quality.
    \item We formulate a theoretical framework and conduct mathematical analyses on our approach, demonstrating its high potential in reducing generalization errors in FSL.
\end{itemize}

\begin{figure*}[ht]
\includegraphics[width=\textwidth]{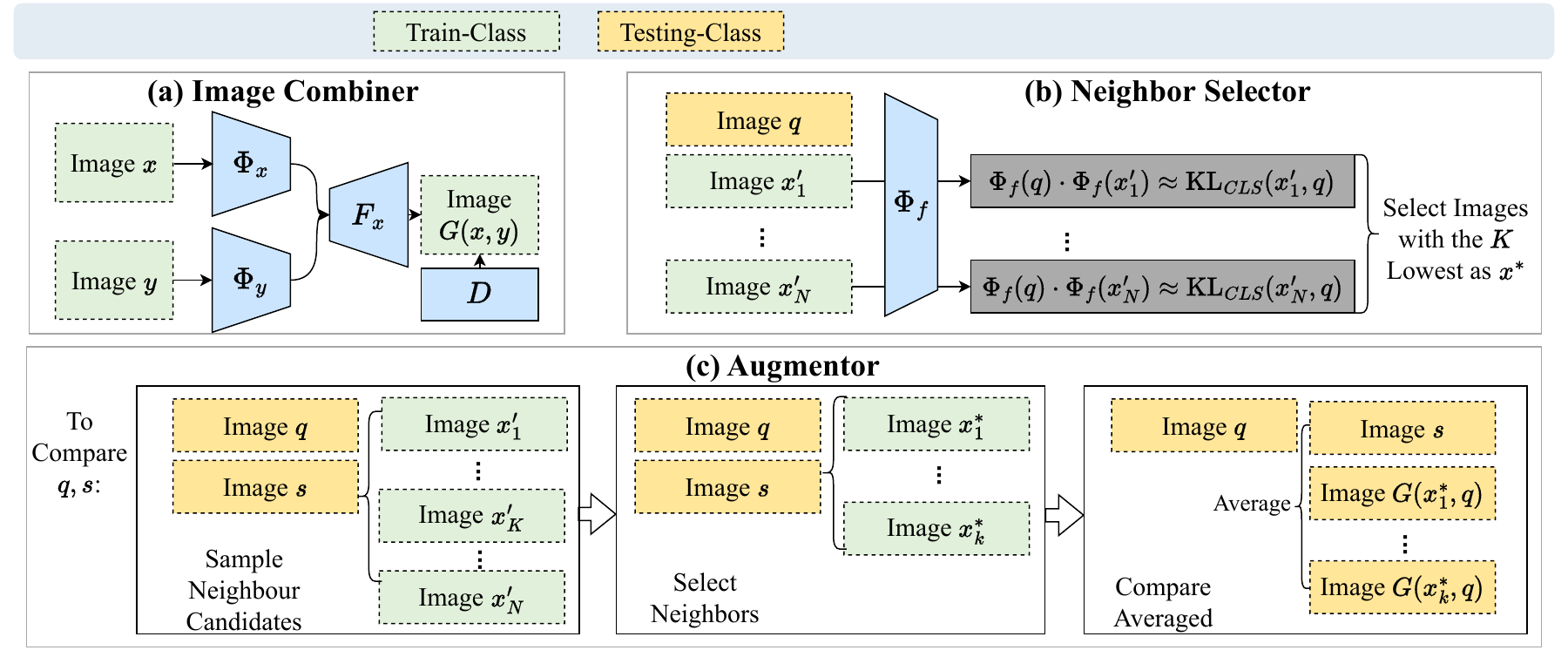}
\caption{Architecture of FSL-Rectifier.}
\label{arch}
\end{figure*}
\section{Related Works}
\textbf{Few-shot learning} sees pioneer works from \citet{fe2003bayesian}, which formulates a bayesian learning framework for quick model adaptation to novel classes. For a broader field of few-shot learning, mainstream algorithms include the meta-learning methodology and metric-learning methodology~\citep{wang2020generalizing, ribeiro2016model, sung2018learning}. For example, meta-learning methods 
\citep{lee2019meta, sun2019meta, rusu2018meta, bertinetto2018metalearning}
like the MAML \citet{finn2017model} learn how to initialize model parameters for training so that the trained parameters can adapt to new unseen tasks quickly. Meanwhile, metric-learning methods like \citet{sung2018learning, snell2017prototypical, ye2020fewshot} aim to learn data representations which are close together as the same class, and far apart as different classes. Related to our work, FSL algorithms include the Matching Network~\citep{https://doi.org/10.48550/arxiv.1606.04080} and the Prototypical Network (ProtoNet)~\citep{snell2017prototypical}, which measure the euclidean distance or cosine similarity between query and support embeddings to identify the most probable categories of the queries. Similar algorithms include FEAT and DeepSets~\citep{ye2020fewshot}, which incorporate a set-to-set transformative layer (i.e., self-attention ~\citep{vaswani2017attention} and Deep Sets function \citep{zaheer2017deep}) to the Matching Network, enhancing the expressiveness of image embeddings. 

\textbf{Image-to-image translation} is a form of generative technique. Deep generative models involve popular architectures like the Variational Auto-Encoder (VAE) and the Generative Adversarial Network (GAN)~\citep{oussidi2018deep, wang2017generative}, while the Adversarial Auto-Encoder combines VAE and GAN. Famous image-to-image translation models include the CycleGAN~\citep{zhu2017unpaired}, which merges two pictures through a symmetric pair of GAN networks. Other related works such as the StyleGAN~\citep{karras2019style}, Coco-Funit/Funit~\citep{saito2020coco, liu2019fewshot} and more~\citep{park2019semantic, wang2018high} can also achieve realistic results in generating new images conditioned on different user requirements. 

\textbf{Data augmentation for FSL} efforts mainly orient towards training data augmentation. Earlier, a ``congealing'' method grafts variations from similar training classes to a separate training class~\citep{wang2020generalizing}. Similar ideas are also seen in FSL works like~\citet{mishra2018generative, verma2018generalized, schwartz2018delta, hariharan2017low, wang2018low, gao2018low}, where the authors augment the training dataset to achieve better generalization among FSL models. 
Another work similar to ours is~\citep{kim2019variational}, where the authors correct the angles of traffic signs through VAE and logo prototype images. 
Although most prior works focus on training-phase augmentation, we only focus on testing-phase augmentation for models that are already trained, which is a novel idea to our best knowledge.

\section{FSL-Rectifier}

\subsection{Preliminary: Few-Shot Learning}
Consider an FSL classifier $h$, which learns from train classes $C^{\text{train}}$ with complete supervision and is then tested on a set of novel classes $C^{\text{test}}$. The training classes and test classes do not overlap (i.e., $C^{\text{train}} \cap C^{\text{test}} = \varnothing$). 
Commonly, during both training and testing, FSL classification tasks follow an \( N \)-way-\( K \)-shot set-up; \( N \) represents the number of classes being classified and \( K \) denotes the number of labelled samples per class. These \( K \times N \) samples, denoted as \( S=\{x_1^{(1)}, x_2^{(1)}, \ldots, x_{K-1}^{(N)}, x_K^{(N)}\} \), are the labelled support set \citep{snell2017prototypical, ye2020fewshot, sung2018learning, wang2020generalizing, dhillon2019baseline}. During \textit{training}, the classifier $h$ is presented with the labelled support set $S$ and unlabelled queries $q$ sub-sampled from the $N$ classes. A classifier $h$ has to learn or predict which categories $q$ belong to by referring to $S$. Here, the $N$ classes are a subset of train classes $C^{\text{train}}$. During \textit{testing}, $h$ has to predict classes of $q$ based on labels of $S$ as well, but both $q$ and $S$ are sampled from the novel test classes $C^{\text{test}}$ \citep{snell2017prototypical, ye2020fewshot, sung2018learning, wang2020generalizing, dhillon2019baseline}.

\subsection{Method}
\paragraph{Overall Pipeline.}
As illustrated in Figure \ref{arch}, our method, FSL-Rectifier, consists of three main components: 1) image combiner, 2) neighbour selector, and 3) augmentor. During stage 1, our goal is to train a generative image combiner that takes in a training sample and a test sample, producing a test-class sample based on the general shape of the training sample. At stage 2, our goal is to train a neighbour selector, which can take in a test sample and a set of candidate training samples. It can return a suitable training sample to be combined with the test sample for generation. The suitable sample is termed \textit{neighbour}. At stage 3, we perform test-time augmentation. For each test sample, we pick its neighbours via the neighbour selector. Then, we produce generated test samples based on neighbours picked and the original test sample, through our image combiner trained at stage 1. For FSL classification with the augmentation setting, we consider the average image embeddings of the original test samples and the generated test samples, instead of just the single original sample.

\paragraph{Image Combiner.}
 Let $x$ denote the image from which we extract the general shape. Let $y$ denote an image from which we extract the class-defining style. The \text{image combiner} $G$ consists of a shape encoder $\Phi_x$, a style encoder $\Phi_y$ and a decoder $F_x$. The style encoder captures the style and appearance of an image, and the shape encoder captures the general shape of another image. The \text{image combiner} essentially produces images $G(x, y)=F_x(\Phi_x(x), \Phi_y(y))$.

During training, we solve the following minimax optimization problem:
\begin{equation}
    \min_{D} \max_{G} \mathcal{L}_{\text{GAN}}(D, G) + k_R\mathcal{L}_
    R(G) + k_{\text{FM}} \mathcal{L}_{\text{FM}}(G) \textrm{.}
\end{equation}
where $D$ is the discriminator in the GAN network, $k_R$ and $ k_{\text{FM}}$ are hyperparemters, $\mathcal{L}_{\text{GAN}}$ is the GAN-loss, $\mathcal{L}_R$ is the reconstruction loss and $\mathcal{L}_{\text{FM}}$ is the feature matching loss \cite{salimans2016improved}. We elaborate on each loss below.

Firstly, the GAN-loss drives the training of both the generator and the discriminator to compete through the following objective function:
\begin{align}
\begin{split}
     \mathcal{L}_{\text{GAN}}(D, G) = \mathbb{E}_x [-\log D(x)] +
     \mathbb{E}_{x,y}[\log(1-D(G(x, y))]\textrm{.}
\end{split}
\end{align}
Here, the discriminator tries to discern between real and images produced by the generator, while encouraging the generation to become the same class as the original test sample class through a classifier $h_d$ coupled with the discriminator encoder $\Phi_d$, where $D = h_d \circ \Phi_d$. $h_d$ is a fully connected layer with output size equal to the number of train classes. At the same time, the generator tries to fool the discriminator.

The reconstruction loss $\mathcal{L}_{R}$ helps generator $G$ generate outputs that resemble the shape of the target images, which are designed as the input images themselves in the loss function:
\begin{equation}
    \mathcal{L}_{R}(G) = \mathbb{E}_x [\|x - G(x,x)\|_1]\textrm{.}
\end{equation}

The feature matching loss helps regularize the training, generating new samples that possess the style of the image $y$: 
\begin{equation}
\begin{split}
      \mathcal{L}_{\text{FM}}(G) = \mathbb{E}_{x, y} \left[\|\Phi_d\left( G(x, y) \right) -  \Phi_d(y) \|_1\right]\textrm{.}
\end{split}
\end{equation}
Here, $\Phi_d$ is the feature extractor, which is obtained from removing the last layer (the classifier layer $h_d$) from the discriminator $D$.

Note that this \text{image combiner}, including the reconstruction loss and feature matching loss, follow prior techniques~\citep{saito2020coco, liu2019few, liu2017unsupervised, park2019semantic, wang2018high, salimans2016improved}. The decoder $F_x$ includes a few adaptive instance normalization (AdaIN) residual blocks~\citep{huang2017arbitrary, liu2019few}. 
Lastly, during training of the image combiner, we only use the train-split of a dataset, leaving the test-split dataset unseen.

\paragraph{Neighbour Selector.}
The \text{image combiner} can at times produce poor results. 
To ensure that good-quality generations are used during the testing phase, we design the \text{neighbour selector} which, when presented with a pool of candidate neighbour images, can return the better candidates for the generation of new test samples. 
Intuitively, a naive way to implement this \text{neighbour selector} is to generate a set of new test samples and select the better based on a measurement of generation quality. However, it can be computationally expensive to generate the actual images for quality assessment. 
Therefore, we aim to ensure the quality while skipping the actual generation.

Reusing the trained \text{image combiner}, we aim to update the discriminator encoder $\Phi_d$ to become $\Phi_f$ such that, for a pair of image samples $\{x, y\}$, $\Phi_f(x)^{\top}\Phi_f(y)$ returns a ``generation quality'' score, which estimates the quality of generation for $G(x, y)$. To achieve this, during training of $\Phi_f$, we feed combinations of train-class image samples and neighbour candidates $\{x, y\}$ and minimize the following objective function $\mathcal{L}_{\textit{KL}}$, defined by:
\begin{equation}\label{kncks}
    \mathbb{E}_{x, y} \left[| \Phi_f(x)^{\top}\Phi_f(y) - \textit{KL}(\sigma(h_d(G(x, y))) \parallel \sigma(h_d(y))) 
      | \right]\textrm{.}
\end{equation} 
Here, $\sigma$ is the softmax function, $h_d(\cdot)$ represents the logits output from the trained and frozen classifier $h_d$ in discriminator $D$. For example, when there are 64 train classes in a dataset, there are 64 logit scores associated with each class, as returned by the classifier $h_d$. We measure the KL-divergence score between the class logit distributions of a potential generation and the target-class sample. When the KL-divergence score is low, the generation is desirable, since the generation $G(x, y)$ tends to be the same in class as its target $y$. We try to train image embeddings that, when multiplied in a pairwise manner, indicate the degree of class difference between the generation and the original sample. Therefore, during actual testing, a candidate neighbour sample $x^*$, whose feature leading to the lowest divergence score when multiplied with the original test sample feature $\Phi_f(y)$, is selected from a pool of neighbour candidates $\{x'\}$: 
\begin{equation}
    x^* = \argmin_{x'}\Phi_f(x')^{\top}\Phi_f(y) \textrm{.}
    \label{neighbour_eqn}
\end{equation}
 As seen in Equation \ref{neighbour_eqn}, we can eliminate the generation of actual new test samples compared with directly using the KL-divergence in Equation \ref{kncks} to measure the generation quality. Meanwhile, the same configuration for the \text{image combiner} is used. We finetune the neighbour selector based on existing model parameters of $D$ learnt from stage 1, keeping the training for neighbour selector simplistic.

\paragraph{Augmentor.}
Suppose we generate $K$ new samples for $N$-way-1-shot classification, and $h \circ \Phi$ is the trained FSL model with encoder $\Phi$ and classifier $h$. The augmentor $\gamma_{K}(y)$ translates one image sample $y$ to $K$ copies of generated images, before considering their average embeddings:

\begin{equation}
\label{feature_avg_eqn}
\begin{split}
    &\gamma_{K}: y \rightarrow 
    \frac{1}{K + 1} \sum\\
    &  (\Phi(G(y,x^*_1))+ \ldots\Phi(G(y,x^*_K)) + \Phi(G(y, y))) \textrm{.}
\end{split}
\end{equation}
Here, $y$ is the test sample and $\{x^*_i\}$ are neighbours. Note that one can tune embedding weights during the embedding summation. Final FSL predictions can be rendered based on classifier $h$:
\begin{equation}
    \hat{y} = h(\gamma_{K}(q)|\{\gamma_{K}(s_1), \ldots \gamma_{K}(s_{n})\})\textrm{.}
\end{equation}
where $q$ represents the query image and $s$ represents each support set image.

\section{Theoretical Insights}
\label{theory}
Here we present how our method with an Support Vector Machine (SVM) \citep{cortes1995support} classifier tends to achieve better performance with one augmentation, a simplified case, from a theoretical perspective. Through Proposition \ref{prop}, we highlight the high probability of outlier (defined by data points with high-value feature norm) reduction. Via Theorem \ref{theorem1}, we associate our method with a usually tighter generalization bound for trained models consisting of a general-architecture encoder and an SVM classifier. This formulation is meaningful for trained FSL models of many configurations, since we do not assume a fixed model encoder architecture (e.g., image feature distribution); for the classifier, it has been formally established that gradient descent iterations on logistic loss and separable datasets converge to hard margin SVM solutions \citep{soudry2024implicitbiasgradientdescent, NIPS2003_0fe47339}. Recently, strong connection between attention-based \citep{vaswani2017attention} classifiers and SVM is also established \citep{tarzanagh2024transformerssupportvectormachines}.

\subsection{Problem Formulation}
We assume that we have a SVM classifier, $h$, trained on pairwise feature differences from the training image embedding differences \{$\mathbf{\Omega} \coloneqq |\Phi(a) - \Phi(b)|; a, b \in \mathcal{D}_\textrm{train}\}$, with labels $l$ defined as:
\begin{equation}
    l(\mathbf{\Omega})= \begin{cases}1 & \text { if } y_a=y_b\textrm{,} \\-1 & \text { otherwise}\textrm{.}\end{cases}
\end{equation}

Note that $\mathbf{\Omega}$ has the same dimension as $\Phi(\cdot)$ under the operation of element-wise absolute difference. Consider an $N$-way-$1$-shot classification task consisting of a query $q$ and support set samples $\{s_1, s_2, \ldots, s_N\}$ during testing. Given pairwise feature differences $\mathbf{\Omega} = \{|\Phi(q) - \Phi(s_1)|, |\Phi(q) - \Phi(s_2)|, \ldots, |\Phi(q) - \Phi(s_N)|\}$, after SVM training, few-shot prediction is rendered as $\operatorname{argmin}_i h(\mathbf{\Omega}_i) \ \text{for} \ 1 \leq i \leq N$. With this set-up, we perform FSL classification via a one-vs-all-classifier \citep{mohri2018foundations}, where each comparison is reduced to a binary classification. In essence, each pairwise feature difference signifies how much two data points differ from each other \citep{bai2024fsl}.

\subsection{Generalization Bound Analysis}

We assume that the test set and augmentation set data are of the same \textit{i.i.d.} data distribution. The test-split feature difference data distribution is defined as $S_1 \coloneqq \{\mathbf{\Omega}: |\Phi(a) - \Phi(b)| ;\ \ a, b \in \mathcal{D}_\textrm{test}\}$. Similarly, $S_2$ is defined for the augmentation test pairs (i.e., $S_2 \coloneqq \{\mathbf{\Omega'}: |\Phi(a') - \Phi(b')| ;\ \ a', b' \in \mathcal{D}_\textrm{aug}\}$). The maximum norm $r$ in $S_1$ is associated with test images $a$ and $b$ (i.e., $r = \max_{S_1} \|\mathbf{\Omega}\| = \||\Phi(a) - \Phi(b)|\|$, $|\Phi(a) - \Phi(b)| \in S_1$).

\begin{prop}
\label{prop}
    When the cardinality of $S_1$ is $v$, we have $\mathbb{P}_{S_1, S_2}(\max_{S_1} \|\mathbf{\Omega}\|>\|\mathbf{\Omega}'\|) = \mathbb{P}_{S_1, S_2}(r>\|\mathbf{\Omega}'\|) = 1 - \frac{1}{2^v}$, $\forall \mathbf{\Omega}' \in S_2$.
\end{prop}

See proof in Appendix Section \textsection \ref{prop1}. This proposition states that the maximum norm $r$ in $S_1$ is larger than any norm in $S_2$ (i.e., $\|\mathbf{\Omega}'\|$) with high probability. Furthermore, the norm of our rectified combination defined as $\|\hat{\mathbf{\Omega}}\|$ can be bounded as follows:
\begin{equation}
\begin{split}
    \|\hat{\mathbf{\Omega}}\| 
    &= \| |(0.5\Phi(a) + 0.5\Phi(a')) - (0.5 \Phi(b) + 0.5 \Phi(b')) |\| \\
    &= \| |0.5(\Phi(a) - \Phi(b)) + 0.5(\Phi(a') - \Phi(b'))|\| \\
    &\leq \|0.5|\Phi(a) - \Phi(b)| + 0.5|\Phi(a') - \Phi(b')|\|\\
    &= 0.5r + 0.5\|\mathbf{\Omega}'\|\textrm{.}
\end{split}
\end{equation}
Therefore, $\|\hat{\mathbf{\Omega}}\|$ can only be larger than $r$ when $\|\mathbf{\Omega}'\| > r$. Since $\|\mathbf{\Omega}'\|$ tends to be smaller than $r$ according to Proposition \ref{prop}, the proposition implies that our method tends to combine the original input in $S_1$ with augmentation in $S_2$ associated with norm smaller than $r$, and as a result, the new norm tends to be smaller than $r$ with high probability. Through our method, the maximum norm tends to reduce regardless of data distribution, especially when $v$ is large.

\begin{defi}[Margin Loss Function]
    For any $\rho>0$, the $\rho$-margin loss is the function $L_\rho: \mathbb{R} \times \mathbb{R} \rightarrow \mathbb{R}_{+}$defined for all $y, y^{\prime} \in \mathbb{R}$ by $L_\rho\left(y, y^{\prime}\right)=\tau_\rho\left(y y^{\prime}\right)$ with
\begin{equation}
\tau_\rho(x)=
\begin{cases}1 &\text{if } x \leq 0 \textrm{,} \\ 1-\frac{x}{\rho} &\text{if } 0 \leq x \leq \rho \textrm{,} \\ 0 &\text{if } \rho \leq x .\end{cases}
\end{equation}
\end{defi}

The parameter $\rho>0$ can be understood as the required confidence margin associated with hypothesis $h$. Suppose we have a margin loss characterized by $\rho$, we have the following generalization bound:
\begin{theorem}
\label{theorem1}
    Let $h=\mathbf{\Omega} \mapsto \mathbf{w} \cdot \mathbf{\Omega}:\|\mathbf{w}\| \leq \Lambda$ and $S \subseteq S_1$. Fix $\rho>0$, then, for any $\delta>0$, with probability at least $1-\delta$ over the choice of a sample $S$ of size $m$, the following holds for $h$ :
$$
R(h) \leq \widehat{R}_{S, \rho}(h)+\underbrace{2 \sqrt{\frac{r^2 \Lambda^2 / \rho^2}{m}}}_{\textrm{Decreases with }r}+\sqrt{\frac{\log \frac{1}{\delta}}{2 m}} .
$$
\end{theorem}
See proof in Appendix Section \textsection \ref{theo_proof}. Note that $R(h) \leq \widehat{R}_{S, \rho}(h)$ is the expected risk with respect to the margin loss characterized by $\rho$, which equals $\mathbb{E}_{S \sim \mathcal{D}_{\textrm{test}}}[\widehat{R}_{S, \rho}(h)]$. The empirical risk $\widehat{R}_{S, \rho}(h)$ equals $\frac{1}{m} \sum_{i=1}^m \tau_\rho\left(l_i h\left(\mathbf{\Omega}_i\right)\right)$, and $l_i \in \{-1, 1\}$ refers to the label associated with $\mathbf{\Omega}_i$. During testing, with augmentation, $r$ tends to decrease with outliers rectified, as implied by Proposition \ref{prop}. Therefore, the second term $2 \sqrt{\frac{r^2 \Lambda^2 / \rho^2}{m}}$ tends to decrease, resulting in a lower generalization error.

\begin{table*}[t]
\centering
\resizebox{\textwidth}{!}{%
    \begin{tabular}{l||
    ccc|ccc}
 \toprule
        &\multicolumn{3}{c|}{Euclidean Distance Classifier on Animals (\%)} &\multicolumn{3}{c}{Cosine Similarity Classifier on Animals (\%)} \\
        \cline{2-7}
        &ProtoNet  & FEAT & DeepSet
         &ProtoNet  & FEAT & DeepSet \\
         \midrule
         \midrule
         No Augmentation
         &{54.64 ± 0.60}  &{46.86 ± 0.59} &{56.48 ± 0.59}
         &{54.56 ± 0.60}  &{46.80 ± 0.59} &55.26 ± 0.59\\
         \hline
         Oracle
         &{80.20 ± 0.56} &{69.96 ± 0.57} &{73.92 ± 0.57}
         &{79.74 ± 0.56} &{69.92 ± 0.57} &74.38 ± 0.57\\
         \hline
         Rotate 
         &{53.68 ± 0.60}  &{50.60 ± 0.59} &{55.24 ± 0.59}
         &{54.68 ± 0.60}  &{50.36 ± 0.59} &53.79 ± 0.59\\
         \hline
         ColorJitter 
         &{52.25 ± 0.60}  &{46.71 ± 0.58} &{54.24 ± 0.58}
         &{52.24 ± 0.60}  &{46.64 ± 0.58} &54.01 ± 0.58\\
         \hline
         Affine 
         &{52.25 ± 0.60}  &{46.07 ± 0.60} &{52.12 ± 0.60}
         &{49.68 ± 0.60}  &{46.68 ± 0.60} &49.41 ± 0.60\\
         \hline
         Mix-Up
         &54.44 ± 0.63 &47.16 ± 0.59 &53.32 ± 0.59
         &54.60 ± 0.63 &47.76 ± 0.59 &53.19 ± 0.59\\
         \hline
         \textbf{FSL-Rectifier (Ours)} 
         &57.90 ± 0.60 (\textit{3.26}$\uparrow)$  &53.28 ± 0.56 (\textit{6.42}$\uparrow)$ &60.18 ± 0.58 (\textit{3.70}$\uparrow)$
         &58.12 ± 0.60 (\textit{3.56}$\uparrow)$  &52.80 ± 0.56 (\textit{6.00}$\uparrow)$ &58.74 ± 0.58 (\textit{3.48}$\uparrow)$\\
 \bottomrule
    \end{tabular}%
    }
    \caption{5-way-1-shot accuracy on Animals datasets, under different trained FSL models.}
    \label{main-result-table}
\end{table*}

\begin{figure*}[t]
\includegraphics[width=\textwidth] {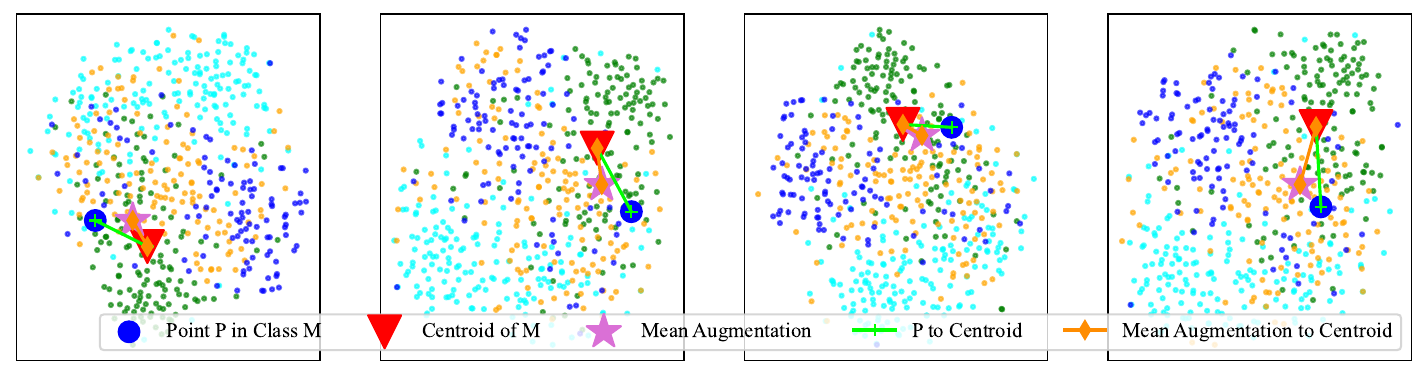}
\caption{TSNE plot indicating that average augmentation of a random point P (\scalebox{1.5}{$\star$}) stay closer to the class centroid ($\blacktriangledown$), compared to the random point P ($\bullet$) on its own. Best viewed in colors.}
\label{tsne}
\end{figure*}
\section{Implementation and Experiments}
\label{imp_sec}
\subsection{Datasets}
In this work, we use the Animals dataset, which is sampled from the ImageNet dataset \citep{ILSVRC15}. The Animals dataset contains carnivorous animal facial images \citep{liu2019few}. The train-test split follows prior works \citep{liu2019few}. For further analysis, we also consider a mammal animal dataset, or the Mammals dataset \citep{asaniczka_2023} consisting of 45 testing classes. 

\subsection{Models and Training}
To obtain trained FSL models used in this work, we use the ProtoNet~\citep{snell2017prototypical}, the FEAT, and the DeepSet~\citep{ye2020fewshot} with either an euclidean-distance or cosine-similarity classifier for experiments. We consider a 4-layer-CNN encoder (Conv4), which we adopt from \citet{ye2020fewshot}. We pretrain the encoder before training it further on different FSL algorithms. We directly employ the pretrained image translator models from \citet{liu2019few, saito2020coco}. When training the neighbour selector, we clone a copy of the trained image combiner. The learning rate is set to 1$\times10^{-4}$, and the maximum amount of training iterations is 10,000. When testing our augmentation against the baseline, the neighbour selector considers 20 candidates for each neighbour selection. 

\subsection{Evaluation Protocol and Results}
We compare our method with other mainstream test-time augmentation techniques \citep{kimura2024understandingtesttimeaugmentation}. Our baselines include: \textbf{1) No Augmentation}: based on original support set and queries without any augmentation; \textbf{2) Oracle}: based on the average of four real test samples for respective support set and queries, which is equivalent to the $5$-way-$4$-shot evaluation benchmark; \textbf{3-5) Rotate/Affine/Color-Jitter}: based on various augmentations adopted from the Pytorch transform functions~\citep{NEURIPS2019_9015}. Rotate is RandomRotation with degree options from $\{0, 180\}$; Affine is RandomAffine with degree options from $\{30, 70\}$, translate options from $\{0.1, 0.3\}$ and scale options from $\{0.5, 0.75\}$; Color-Jitter is ColorJitter with brightness, contrast and saturation as 0.2, hue as 0.1.
Besides, we perform \textbf{6) Mix-Up} defined as the pixel-level average between the original test sample and its neighbour. For our method, the combined weight of all augmentations and that of the original samples are both set to 0.5. To avoid repeated inference, we save augmentation for each original image, where each generation based on 20 neighbour candidates takes around 0.47 seconds to be generated via one NVIDIA RTX A5000. We pass 25,000 5-way-1-shot queries to test a trained FSL model. 5-way-1-shot accuracy and 95\% confidence intervals are reported. Table \ref{main-result-table} consists of experimental results of our proposed method over the Animals dataset. Based on the results, our method can improve the original FSL model performance by around 4\% on average, which is more effective compared to other common test-time augmentation techniques. Note that the FEAT is not trained till full convergence under the given set of hyper-parameters, and Table \ref{mammal-result} includes additional results over the Mammals dataset without the neighbour selector.

 \begin{figure*}[t]
    \includegraphics[width=\textwidth] {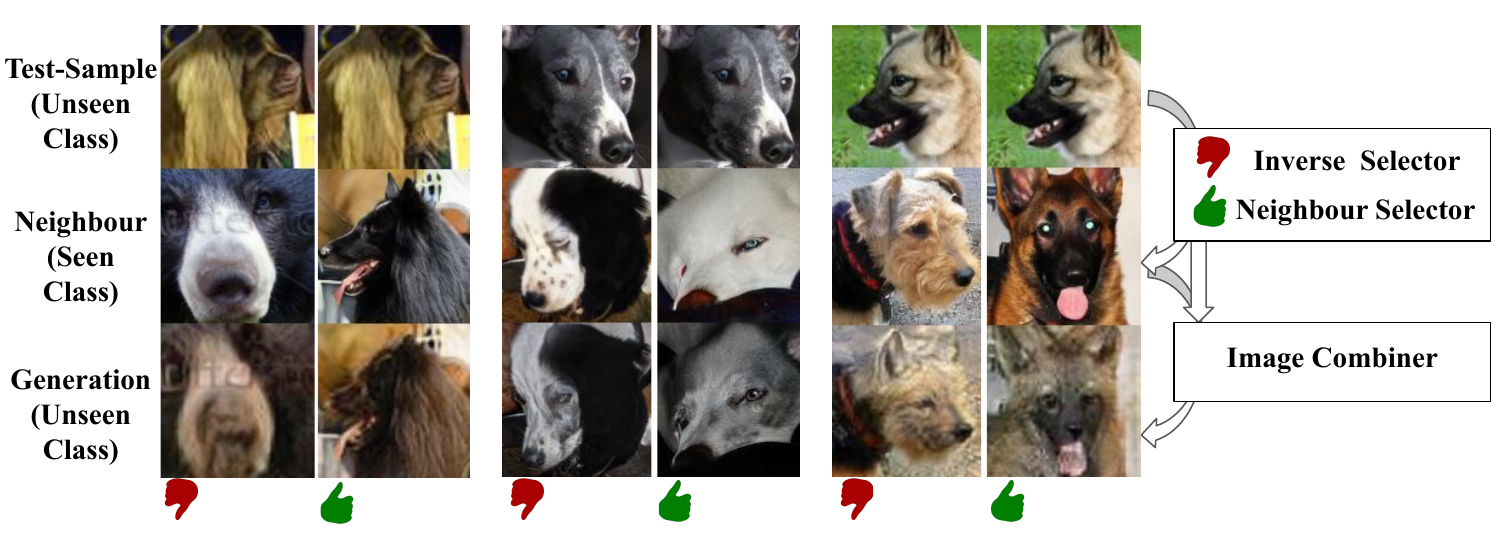}
    \caption{Illustration of the effect of neighbour selector. The first row are the original test samples, the second row are neighbours picked by either neighbour selector or inverse neighbour selector. The third row are new test samples generated by the \text{image combiner} based on neighbours picked by different neighbour selectors.}
    \label{selector}
\end{figure*}
\section{Further Discussions and Ablation Studies}
\label{aug_qty_sec}

\paragraph{Near to Centroids and Reduced Norm?}
Since centroids are defined by the averaged representation of a class, we speculate that our test sample features become closer to class centroids when we consider the average of more same-class copies instead of just one copy. We randomly sample images from 4 classes and visualize the TSNE \citep{vanDerMaaten2008} plot of the sampled images. Figure \ref{tsne} summarises our visualization. The blue point $P$ is a randomly sampled point in class $M$, and we always observe that its rectified version (i.e., averaged with augmentations) comes closer to the centroid. Intuitively, when the points to be classified become nearer to the centroids, the performance degradation caused by the outlier effect is reduced.

Moreover, as discussed in Section \ref{theory}, the maximum norm of pairwise feature difference tends to reduce through our method. We randomly sample 4,000 image feature pairs and measure the maximum pairwise norm (i.e., $\mathbf{\Omega}$), calculated before and after 3 copies of augmentation. We repeat the sampling and calculation 15 times. Although the augmented copies are of a different data distribution, where a higher maximum norm of difference is expected, we still observe a reduced maximum norm for our method over all trials. As indicated in Table \ref{max_norm_table}, the average maximum norm reduces from 21.37 to 19.08. The reduction is consistent with our theoretical analyses.
\begin{table}[ht]
\centering
    \begin{tabular}{l||cc
    }
    \toprule
    &Mean &STD\\
         \toprule
         \toprule
                 Norm Before &21.37 &0.81\\
         \hline
                 Norm After &19.08 &0.57\\
  \bottomrule
    \end{tabular}
    \caption{Maximum norm of randomly sampled absolute-valued feature difference, before and after rectification.}
    \label{max_norm_table}
\end{table}

\paragraph{Without Neighbour Selector?}
To study the importance of our neighbour selector, we first visualize the neighbours and generations rendered by an \textit{inverse neighbour selector} and the original neighbour selector. To recap, the neighbour selector sorts a set of candidate neighbours from good to bad. The inverse neighbour selector is the same but with reversed sorting. The contrast is presented in Figure \ref{selector}, demonstrating better generation returned by the neighbour selector. Meanwhile, as indicated in Table \ref{tab:selector}, the inverse neighbour selector downgrades the ProtoNet performance on Animals dataset from 54.64\% to around 51\%. The random neighbour selector can only improve the original FSL model accuracy score to 56.70\%. In this study, we set the number of \text{neighbour} candidates to consider for each generation to 20. 
\begin{table}[h]
\centering
    \begin{tabular}{l||c
    }
    \toprule
    &5-Way Accuracy (\%)\\
         \toprule
         \toprule
                 No Augmentation &54.64 \\   
         \hline
                 Neighbour Selector &57.90 \\
         \hline
                 Inverse Neighbour Selector &51.08 \\
         \hline
                 Random Neighbour Selector &56.70 \\
  \bottomrule
    \end{tabular}
    \caption{Effect of neighbour selector and its ablative variants.}
    \label{tab:selector}
\end{table}

\paragraph{How Much Augmentation?}
\begin{table}
    \centering
\resizebox{\columnwidth}{!}{%
    \begin{tabular}{c||
    c|
    c|
    c|
    c
    }
    \toprule
         \multicolumn{1}{c||}{Augmentation Copies} &\multicolumn{1}{c|}{(Support) 0}&  \multicolumn{1}{c|}{1}&  \multicolumn{1}{c|}{2}
         &\multicolumn{1}{c}{3}
         \\
         \toprule
         \toprule
                  (Query) 0
                  &\textbf{54.64} &56.60 &57.31 &57.61 \\
         \hline
                  1 
                  &53.46 &54.24 &55.95 &55.63 \\
         \hline
                  2 
                  &54.27 &56.01 &56.47 &57.10\\
         \hline
                  3 
                  &54.30 &56.30 &57.18 &\textbf{57.90}\\
  \bottomrule
    \end{tabular}%
    }
    \caption{How sizes of augmentation affect euclidean-distance-based ProtoNet accuracy.}
    \label{query-table2}
\end{table}
To study how many samples to augment, we augment each input query with less than or equal to 3 additional copies. Table \ref{query-table2} summarises our results on the trained ProtoNet model of a euclidean-distance classifier. We observe that the best result (i.e., 57.90\%) is achieved by 3 augmentation copies for both the support set samples and the queries. Meanwhile, it is important to augment the queries when we augment the support samples, otherwise the support augmentation may backfire (e.g., Degradation from 54.64\% to 53.46\% when there is 1 query augmentation but no support augmentation.). This phenomenon is expected, since the augmentation copies have different data distributions compared to the original images (e.g., different image resolutions). Thus, the augmentation sizes for support-query pairs should match to achieve consistency and optimality.

\section{Conclusions}
On the whole, we propose a method to improve the performance of the trained FSL model through test-time augmentation, during which the \text{image combiner} converts suitable training samples, picked by the neighbour selector, to test classes. The averaged representation becomes more typical especially when compared to outliers, thus improving a trained FSL model without costing extra dataset or training. This approach explores the alternative possibility that differs from previous training augmentation techniques, and our experiments and theoretical analyses demonstrate its feasibility and effectiveness.

\appendix
\section*{Appendices}
\section{Additional Results on Mammals}
Table \ref{mammal-result} summarises additional results over the Mammals dataset without the neighbour selector.
\begin{table}[ht]
\centering
\resizebox{\columnwidth}{!}{%
    \begin{tabular}{l||cc}
 \toprule
        &\multicolumn{2}{|c}{Cosine Similarity Classifier on Mammals (\%)}\\
        \cline{2-3}
         &ProtoNet & DeepSet \\
         \midrule
         \midrule
         No Augmentation
         &{43.07 ± 0.55} &{44.47 ± 0.55} \\
         \hline
         Oracle
         &{76.02 ± 0.50} &{76.70 ± 0.50} \\
         \hline
         Rotate 
         &{42.04 ± 0.55} &{43.04 ± 0.54} \\
         \hline
         ColorJitter 
         &{42.03 ± 0.54} &{42.36 ± 0.54} \\
         \hline
         Affine 
         &{42.39 ± 0.53} &{42.66 ± 0.53}\\
         \hline
         Mix-Up
         &41.76 ± 0.56 &41.32 ± 0.52 \\
         \hline
         \textbf{FSL-Rectifier (Ours)} 
         &44.07 ± 0.55 &46.20 ± 0.54\\
 \bottomrule
    \end{tabular}%
    }
    \caption{5-way-1-shot accuracy on Mammals datasets, under different trained FSL models.}
    \label{mammal-result}
\end{table}

\section{Proof of Proposition 1.}
\label{prop1}
Given that $S_1$ and $S_2$ are of the same \textit{i.i.d.} distribution, via the law of symmetry, we have $\mathbb{P}_{S_1, S_2}(\|{\mathbf{\Omega}}\| \leq \|\mathbf{\Omega}'\|) = \frac{1}{2}$, $\forall \ \mathbf{\Omega}$ and $\mathbf{\Omega}'$ in $S_1$ and $S_2$ respectively. Consider a set $S_1$ consisting of $v$ samples. We have:
\begin{equation}
\begin{split}
    \mathbb{P}_{S_1, S_2}&(\max_{S}\|{\mathbf{\Omega}}\|>\|\mathbf{\Omega}'\|) \\
    &= 1 - \mathbb{P}_{S_1, S_2}(\max_{S}\|{\mathbf{\Omega}}\| \leq \|\mathbf{\Omega}'\|)\\
    &= 1 - (\mathbb{P}_{S_1, S_2}(\|{\mathbf{\Omega}_1}\| \leq \|\mathbf{\Omega}'\|) \times\\ &\quad \quad \mathbb{P}_{S_1, S_2}(\|{\mathbf{\Omega}_2}\| \leq \|\mathbf{\Omega}'\|) \ldots \times \\
    &\quad \quad  \mathbb{P}_{S_1, S_2}(\|{\mathbf{\Omega}_v}\| \leq \|\mathbf{\Omega}'\|))\\
    &= 1 - (\mathbb{P}_{S_1, S_2}(\|{\mathbf{\Omega}}\| \leq \|\mathbf{\Omega}'\|))^v\\
    &= 1 - \frac{1}{2^v}\textrm{.}
\end{split}
\end{equation}
\qed

\section{Proof of Theorem 1.}
\label{theo_proof}
Let $\mathcal{H}$ be our hypothesis set. Consider the family of functions taking values in $[0,1]$ :
\begin{equation}
\tilde{\mathcal{H}}=\left\{\tau_\rho \circ l h: h \in {\mathcal{H}}\right\}\textrm{.}
\end{equation}

By the Rademacher Complexity Generalization Bound in \citet{mohri2018foundations}, with probability at least $1-\delta$, for all $g \in \tilde{\mathcal{H}}$,
\begin{equation}
\mathbb{E}[g(\mathbf{\Omega})] \leq \frac{1}{m} \sum_{i=1}^m g\left(\mathbf{\Omega}_i\right)+2 \Re_m(\tilde{\mathcal{H}})+\sqrt{\frac{\log \frac{1}{\delta}}{2 m}}\textrm{.}
\end{equation}
Note that $g(\mathbf{\Omega}) = \tau_\rho \circ l h(\mathbf{\Omega})$,  $\Re_m(\tilde{\mathcal{H}}) = \frac{1}{m} \underset{\sigma, S}{\mathbb{E}}\left[\sup _{h \in \mathcal{H}} \sum_{i=1}^m \sigma_i \tau_\rho(l_i h\left(\mathbf{\Omega}_i\right)\right)]$, $m$ is the cardinality of sample set $S$ drawn from $S_1$ (i.e., absolute feature differences associated with original test samples) and $\sigma$ takes a random value from $\{-1, 1\}$ with uniform probability. 

$\Re_m(\tilde{\mathcal{H}})$ can be rewritten as:
\begin{equation}
\begin{split}
    \Re_m(\tilde{\mathcal{H}}) &= \frac{1}{m} \underset{\sigma, S}{\mathbb{E}}\left[\sup _{h \in \mathcal{H}} \sum_{i=1}^m \sigma_i \tau_\rho(l_i h\left(\mathbf{\Omega}_i\right)) \right]\\
    &=\frac{1}{m} \underset{\sigma, S}{\mathbb{E}}\left[\sup_{h \in \mathcal{H}} \sum_{i=1}^m \sigma_i \tau_\rho(h\left(\mathbf{\Omega}_i\right))\right]=
\Re_m(\tau_\rho \circ \mathcal{H})\textrm{.}
\end{split}
\end{equation}

Let $\widehat{R}_{S, \rho}(h)$ equal $\frac{1}{m} \sum_{i=1}^m g\left(\mathbf{\Omega}_i\right)$. For all $h \in \mathcal{H}$:
\begin{equation}
\mathbb{E}\left[\tau_\rho(l h(\mathbf{\Omega}))\right] \leq \widehat{R}_{S, \rho}(h)+2 \Re_m\left(\tau_\rho \circ \mathcal{H}\right)+\sqrt{\frac{\log \frac{1}{\delta}}{2 m}}\textrm{.}
\end{equation}

Since $1_{u \leq 0} \leq \tau_\rho(u)$ for all $u \in \mathbb{R}$, we have $R(h)=\mathbb{E}\left[1_{l h(\mathbf{\Omega}) \leq 0}\right] \leq \mathbb{E}\left[\tau_\rho(l h(\mathbf{\Omega}))\right]$, thus:
\begin{equation}
R(h) \leq \widehat{R}_{S, \rho}(h)+2 \Re_m\left(\tau_\rho \circ \mathcal{H}\right)+\sqrt{\frac{\log \frac{1}{\delta}}{2 m}}\textrm{.}
\end{equation}

Since $\tau_\rho$ is $1 / \rho$-Lipschitz, by the Talagrand's Lemma in \citet{mohri2018foundations}, we have $\Re_m\left(\tau_\rho \circ \mathcal{H}\right) \leq \frac{1}{\rho} \Re_m(\mathcal{H})$.
Therefore:
\begin{equation}
    \begin{aligned}
    R(h) & \leq \widehat{R}_{S, \rho}(h)+\frac{2}{\rho} \Re_m(\mathcal{H})+\sqrt{\frac{\log \frac{1}{\delta}}{2 m}}\textrm{.}
    \end{aligned}
\end{equation}

    Note that $S \subseteq\{\mathbf{\Omega}:\|\mathbf{\Omega}\| \leq r\}$ is a sample set of size $m$ drawn from $S_1$ and $\mathcal{H}=\{\mathbf{\Omega} \mapsto$ $\mathbf{w} \cdot \mathbf{\Omega}:\|\mathbf{w}\| \leq \Lambda\}$. Then, the empirical Rademacher complexity of $\mathcal{H}$ can be bounded as:
\begin{equation}
\begin{split}
\Re_m(\mathcal{H}) & =\frac{1}{m} \underset{\sigma, S}{\mathbb{E}}\left[\sup _{\|\mathbf{w}\| \leq \Lambda} \sum_{i=1}^m \sigma_i \mathbf{w} \cdot \mathbf{\Omega}_i\right]\\
&=\frac{1}{m} \underset{\sigma, S}{\mathbb{E}}\left[\sup _{\|\mathbf{w}\| \leq \Lambda} \mathbf{w} \cdot \sum_{i=1}^m \sigma_i \mathbf{\Omega}_i\right] \\
& \leq \frac{\Lambda}{m} \underset{\sigma, S}{\mathbb{E}}\left[\left\|\sum_{i=1}^m \sigma_i \mathbf{\Omega}_i\right\|\right] \\
& \leq \frac{\Lambda}{m}\left[\underset{\sigma, S}{\mathbb{E}}\left[\left\|\sum_{i=1}^m \sigma_i \mathbf{\Omega}_i\right\|^2\right]\right]^{\frac{1}{2}} \\
& =\frac{\Lambda}{m}\left[\underset{\sigma, S}{\mathbb{E}}\left[\sum_{i, j=1}^m \sigma_i \sigma_j\left(\mathbf{\Omega}_i \cdot \mathbf{\Omega}_j\right)\right]\right]^{\frac{1}{2}} \\
& \leq \frac{\Lambda}{m}\left[\underset{S}{\mathbb{E}}\left[\sum_{i=1}^m\left\|\mathbf{\Omega}_i\right\|^2\right]  \right]^{\frac{1}{2}}\leq \frac{\Lambda \sqrt{m r^2}}{m}=\sqrt{\frac{r^2 \Lambda^2}{m}}\textrm{.}
\end{split}
\end{equation}

The first inequality uses the Cauchy-Schwarz inequality and the bound on $\|\mathbf{w}\|$, the second inequality uses the Jensen's inequality, the third inequality uses $\mathbb{E}\left[\sigma_i \sigma_j\right]=\mathbb{E}\left[\sigma_i\right] \mathbb{E}\left[\sigma_j\right]=$ 0 for $i \neq j$, and the last inequality uses $\left\|\mathbf{\Omega}_i\right\| \leq r$. \qed

\section*{Acknowledgements}
This research is supported in part by the Ministry of Education, Singapore, under its MOE AcRF TIER 3 Grant (MOE-MOET32022-0001). The authors would like to thank Jiangwei Chen, Xiangming Gu, Dexter Neo, Miao Xiong, Junfeng Hu, Xiangyu Peng and Haonan Wang for their feedback.

\bibliography{aaai25}
\end{document}